%% file: GSC.tex
\documentclass[runningheads,a4paper]{llncs}

\usepackage{url}

\usepackage{times}

\usepackage{graphicx} 

\usepackage{textpos}
\usepackage{amssymb} 
\usepackage{amsmath}
\usepackage{bbm}
\usepackage{dsfont}
\usepackage{graphicx}
\usepackage{color}
\usepackage{prelim2e}
\graphicspath{{../pics/}}
\definecolor{brightblue}{rgb}{0.0,0.0,1.0}
\definecolor{darkblue}{rgb}{0.0,0.0,0.5}
\definecolor{darkblack}{rgb}{0.0,0.0,0.0}

\newcommand{\comment}[1]{}

\urldef{\mailsa}\path|{luecke, sheikh}@fias.uni-frankfurt.de|

\newcommand{\One}{\hbox{$1\hskip -1.2pt\vrule depth 0pt height 1.6ex width 0.7pt\vrule depth 0pt height 0.3pt width 0.12em$}}
\newcommand{\disT}{\textstyle}

\newcommand{\FF}{{\cal F}}
\newcommand{\sVec}{\vec{s}}

\newcommand{\zVec}{\vec{z}}

\newcommand{\yVec}{\vec{y}}
\newcommand{\yVecN}{\vec{y}^{\,(n)}}
\newcommand{\piVec}{\vec{\pi}}

\newcommand{\WVec}{\vec{W}}
\newcommand{\Wt}{\tilde{W}}

\newcommand{\kappaVec}{\vec{\kappa}}

\newcommand{\kappaVecN}{\kappaVec^{(n)}}

\newcommand{\E}[1]{\big\langle{}#1\big\rangle}

\newcommand{\RRR}{\mathbbm{R}}

\newcommand{\ThetaOld}{\Theta^{\mathrm{old}}}
\newcommand{\NGauss}{{\cal N}}

\newcommand{\sig}{\sigma}
\newcommand{\GSCParamsList}{(W,\Sigmad,\piVec)}
\newcommand{\dz}{\mathrm{d}\zVec}

\newcommand{\trace}{\mathrm{Tr}}
\newcommand{\TT}{\mathrm{T}}
\newcommand{\refp}[1]{(\ref{#1})}
\newcommand{\ssb}{\hspace{-2mm}}

\newcommand{\Sigmad}{\Sigma}
\newcommand{\Sigmah}{\One_{H}}

\begin{document}

\mainmatter  

\title{Closed-form EM for Sparse Coding\\
And its Application to Source Separation
}

\titlerunning{Closed-form EM for Sparse Coding and its Application to Source Separation}

%
%
\author{J\"org L\"ucke$^{\star }$ \and Abdul-Saboor Sheikh%
\thanks{joint first authorship}%
}
\authorrunning{J\"org L\"ucke and Abdul-Saboor Sheikh}

\institute{FIAS, Goethe-University Frankfurt, 60438 Frankfurt, Germany
\mailsa\\
}

%
%

\toctitle{Closed-form EM for Sparse Coding and its Application to Source Separation}
\tocauthor{J\"org L\"ucke \& Abdul-Saboor Sheikh}
\maketitle

\begin{abstract}
  We define and discuss the first sparse coding algorithm based on
  closed-form EM updates and continuous latent variables.  The
  underlying generative model consists of a standard
  `spike-and-slab' prior and a Gaussian noise model.
  Closed-form solutions for E- and M-step equations are derived by
  generalizing probabilistic PCA.
  The resulting EM algorithm can take all modes of a potentially
  multi-modal posterior into account.  The computational cost of the
  algorithm scales exponentially with the number of hidden dimensions.
  However, with current computational resources, it is still possible 
  to efficiently learn model parameters for medium-scale problems.
  Thus the model can be applied to the typical range of source separation tasks.
  In numerical experiments on artificial data we verify likelihood
  maximization and show that the derived algorithm recovers the sparse
  directions of standard sparse coding distributions. On source
  separation benchmarks comprised of realistic data we show that the
  algorithm is competitive with other recent methods.
\end{abstract} 

\begin{textblock}{9}(0.0,5.5)
{\em Preprint. Final version to appear in Proc. LVA/ICA, LNCS pp. 213-221, 2012.}
\end{textblock}

\section{Introduction}
\label{SecIntro}
Probabilistic generative models are a standard approach to model data
distributions and to infer instructive information about the
data generating process. 
%
%
Methods like principle component analysis, factor analysis, or
sparse coding (SC) (e.g., \cite{OlshausenField1996}) have all been
formulated in the form of probabilistic generative models.
Moreover, independent component analysis (ICA), which
is a very popular approach to blind source separation, 
can also be recovered from sparse coding in the 
limit of zero observation noise (e.g., \cite{DayanAbbott2001}).

%
%
%
A standard procedure to optimize parameters in generative models is 
the application of Expectation Maximization (EM)
(e.g., \cite{NealHinton1998}).
However, for many generative models the optimization using EM is
analytically intractable. For stationary data only the most elementary models such
as mixture models and factor analysis (which contains probabilistic
PCA as special case) have closed-form solutions for E- and M-step
equations. EM for more elaborate models requires approximations. In
particular, sparse coding models (\cite{OlshausenField1996,LeeEtAl2007,Seeger2008} and many
more) require approximations because
integrals over the latent variables do not have closed-form solutions.

In this work we study a generative model that combines the Gaussian
prior of probabilistic PCA (p-PCA) with a binary prior distribution.
Distributions combining binary and continuous parts have been
discussed and used as priors before (e.g., \cite{MitchellBeauchamp1988}) and are commonly referred to as
`spike-and-slab' distributions. Also sparse coding variants with
spike-and-slab distributions have been studied previously (compare
\cite{West2003,KnowlesGhahramani2007,TehGorurGhar2007,PaisleyLawrence2009,KnowlesGhahramani2010,MohamedEtAl2010}).
However, in this work we show that combining binary and Gaussian latents
maintains the p-PCA property of having a closed-form solution for EM
optimization. We can, therefore, derive an algorithm that uses exact posteriors
with potentially many modes to update model parameters.
%
\begin{figure}
\vspace{-8mm}
\begin{minipage}[b]{7.45cm}
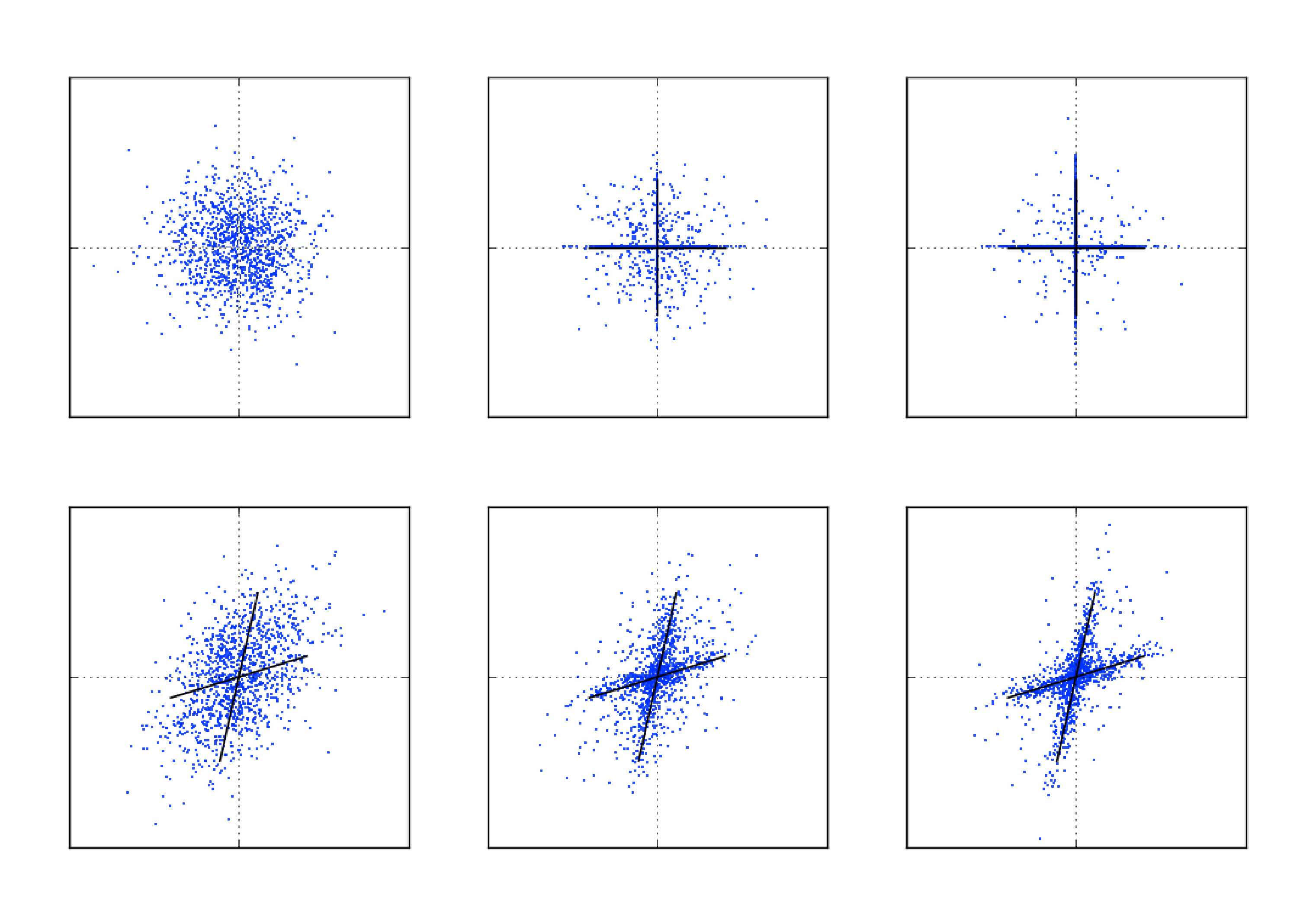
\label{FigGSCDistributions}
\end{minipage}\hfill
\raisebox{2.75cm}{
\begin{minipage}{4.75cm}
\caption{\small Distributions generated by the GSC generative model.
The left column shows the distributions generated for $\pi_h=1$ for all $h$.
In this case the model generates p-PCA distributions. The middle column
shows
an intermediate value of $\pi_h$. The generated distributions are not
Gaussians
anymore but have a slight star shape. The right column shows
distributions for
small values of $\pi_h$. The generated distributions have a salient star
shape
similar to standard sparse coding distributions.}
\end{minipage} }
\vspace{-10mm}
\end{figure}
\section{The Gaussian Sparse Coding (GSC) model} 
\label{SecGSCmodel}
Let us first consider a pair of $H$--dimensional i.i.d. latent vectors, a continuous $\zVec${}$\in${}$\RRR^H$ 
and a binary $\sVec${}$\in${}$\{0,1\}^H$ with:
\begin{eqnarray}
\ssb{}p(\sVec\,|\Theta) &=& \prod_{h=1}^{H}\pi_h^{s_h}\,(1-\pi_h)^{1-s_h} = \small{\mathrm{Bernoulli}}(\sVec;\piVec)\label{EqnPriorS} \mbox{\ \ and\ \  }
p(\zVec\,|\Theta) = \NGauss(\zVec;\,\vec{0},\Sigmah)\label{EqnPriorZ},
\end{eqnarray}
%
where $\pi_h$ parameterizes the probability of non-zero entries.
After generation, both hidden vectors are combined using a pointwise
multiplication operator: 
i.e., $(\sVec\odot\zVec)_h = s_h\,z_h$ for all $h$.
%
%
%
%
The resulting hidden random variable is a vector
of continuous values and zeroes, and it follows a 
`spike-and-slab' distribution.
%
%
Given a hidden vector (which we will denote
by $\sVec\odot\zVec$) 
, we generate a $D$--dimensional observation
$\yVec\in\RRR^{D}$ by linearly combining a set of basis functions $W$ and 
adding Gaussian noise:
\begin{eqnarray}
\ssb\ssb{}p(\yVec\,|\,\sVec,\zVec,\Theta) &=& \NGauss(\yVec;\,W(\sVec\odot\zVec),\Sigmad),
\label{EqnNoise}
\end{eqnarray}
%
where $W\in\RRR^{D\times{}H}$ is the matrix containing the basis
functions $\WVec_h$ as columns, and $\Sigmad\in\RRR^{D\times{}D}$ is a covariance matrix
parameterizing the data noise.
The latents' priors \refp{EqnPriorS} together with their pointwise combination 
and the noise distribution \refp{EqnNoise} define the generative model under
consideration. As a special case, the model contains probabilistic PCA (or factor analysis).
This can easily be seen by setting all $\pi_h$ equal to one.

%
The model \refp{EqnPriorS} to \refp{EqnNoise} is capable of generating
a broad range of distributions including sparse coding like
distributions. This is illustrated in Fig.\,\ref{FigGSCDistributions}
where the parameters $\pi_h$ allow for
continuously changing PCA-like to a SC-like distribution.
  
While the generative model itself has been studied previously
\cite{West2003,TehGorurGhar2007,PaisleyLawrence2009,KnowlesGhahramani2010},
we will show that a closed-form EM algorithm can be derived, 
which can be applied to blind source separation tasks. We will refer to the
generative model \refp{EqnPriorS} to \refp{EqnNoise} as the {\em
  Gaussian Sparse Coding} (GSC) model in order to stress that a
specific spike-and-slab prior (Gaussian slab) in conjunction with a
Gaussian noise model is used. The GSC model is thus an instance of
the spike-and-slab sparse coding model (or alternatively known 
{\em sparse factor analysis} models;  
 see e.g., \cite{West2003,TehGorurGhar2007,PaisleyLawrence2009,KnowlesGhahramani2010}).
\subsection{Expectation Maximization (EM) for Parameter Optimization}
\label{SecEM}
Consider a set of $N$ independent data points \mbox{$\{\yVecN\}_{n=1,\ldots,N}$}
with $\yVecN\in\RRR^D$.  For these data we seek
parameters $\Theta=\GSCParamsList$ that maximize the data likelihood
${\cal L}=\prod_{n=1}^N{}p(\yVecN\,|\,\Theta)$ under the GSC generative model.
We employ Expectation Maximization (EM) algorithm for parameter optimization.
The EM algorithm \cite{NealHinton1998} optimizes the data 
likelihood w.r.t.\ the parameters $\Theta$ by iteratively maximizing the free-energy given by:
%
%
\begin{eqnarray}
\hspace{-2mm} \FF(\ThetaOld,\Theta) &=& \disT
    \sum\limits_{n=1}^{N} \sum\limits_{\sVec}\int\limits_{\zVec}\ p(\sVec,\zVec\,|\,\yVecN,\ThetaOld) 
    \disT  \Big[\log\big( p(\yVecN\,|\,\sVec,\zVec,\Theta)\big)+ \log \big( p(\sVec\,|\,\Theta) \big) \nonumber\\
\hspace{-2mm} &&\hspace{47mm}\disT  + \log \big( p(\zVec\,|\,\Theta) \big) \Big]\,\dz\,+\,H(\ThetaOld)\,,
\label{EqnFreeEnergy}
\end{eqnarray}
where $H(\ThetaOld)$ is an entropy term only depending on parameter
values held fixed during the optimization of $\FF$ w.r.t.\ $\Theta$.
Note that integration over the hidden space involves an integral over
the continuous part and a sum over the binary part.

Optimizing the free-energy consists of two steps: given the current
parameters $\ThetaOld$ the posterior probability is computed in the
E-step; and given the posterior, $\FF(\ThetaOld,\Theta)$ is maximized
w.r.t.\ $\Theta$ in the M-step. Iteratively applying E- and M-steps
locally maximizes the data likelihood.
%
%
%
%
%
%
%
%
%
%
%
%
%
%
%
%
%
\\[2mm]
{\bf M-step parameter updates:} Let us first consider the
maximization of the free-energy in the M-step before considering
expectation values w.r.t.\ to the posterior in the E-step. Given a generative
model, conditions for a maximum free-energy are canonically derived by
setting the derivatives of $\FF(\ThetaOld,\Theta)$ w.r.t.\ the second
argument to zero. For the GSC model we obtain the following parameter updates:
%
%
%
\small
\begin{eqnarray}
W &=& \big( \sum_{n=1}^{N} \yVecN \E{\sVec\odot\zVec}^{\TT}_n \big) 
             \big( \sum_{n=1}^{N} \E{(\sVec\odot\zVec)(\sVec\odot\zVec)^{\TT}}_n \big)^{-1},\label{EqnMStepW}\\
\Sigmad &=& \frac{1}{N}\sum_{n=1}^{N}\Big[ \yVecN(\yVecN)^{\TT}
                    - 2 \big(W \E{\sVec\odot\zVec}_n\big)(\yVecN)^{\TT} + W\big(\E{(\sVec\odot\zVec)(\sVec\odot\zVec)^{\TT}}_nW^{\TT}\big)\Big]\label{EqnMStepSigma} \nonumber\\
& &\hspace{-7mm}\mbox{and}\hspace{2mm} \piVec = \frac{1}{N}\sum_{n=1}^{N}\E{\sVec}_n,  \label{EqnMStepPi}
%
%
\hspace{2mm}\mbox{where}\hspace{2mm}\E{f(\sVec,\zVec)}_n = \sum_{\sVec}\int_{\zVec}\,p(\sVec,\zVec\,|\,\yVecN,\ThetaOld)\ f(\sVec,\zVec)\,\dz.\label{EqPstrXpt}
\end{eqnarray}
\normalsize
%
%
%
Equations \refp{EqnMStepW} to \refp{EqnMStepPi} define a new set
of parameter values $\Theta=\GSCParamsList$ given the current
values $\ThetaOld$. These 'old' parameters are only used to 
compute the sufficient statistics 
$\E{\sVec}_n$, $\E{\sVec\odot\zVec}_n$
and $\E{(\sVec\odot\zVec)(\sVec\odot\zVec)^{\TT}}_n$ of the model. 
\\[2mm]
{\bf Expectation Values:} Although the derivation of M-step equations can be analytically
intricate, it is the E-step that, for most generative models, poses
the major challenge.
Source of the problems involved are analytically intractable integrals
required for posterior distributions and for expectation values
w.r.t.\ the posterior. The true posterior is therefore often replaced
by an approximate distribution (see, e.g.,\cite{Bishop2006,Seeger2008}) or in
the form of factored variational distributions \cite{JordanEtAl1999,Jaakkola2001}.
The most frequently used approximation is the
maximum-a-posterior (MAP) estimate (see, e.g., \cite{OlshausenField1996,LeeEtAl2007}) 
which replaces the
true posterior by a delta-function around the posterior's maximum
value. Alternatively, analytically intractable expectation values are
often approximated using sampling approaches. 
%
%
Using approximations always implies, however, that many analytical
properties of exact EM are not maintained. Approximate EM iterations
may, for instance, decrease the likelihood or may not recover (local
or global) likelihood optima in many cases. 
There are nevertheless, a limited number of models with exact EM solutions; e.g., 
mixture models such as the mixture-of-Gaussians, p-PCA or factor analysis etc. 
Our novel work here extends the set of known models with exact EM solutions.
By following along the same lines as for the p-PCA derivations, we maintain 
in our E-step the analytical tractability of computing expectation values 
w.r.t. the posterior of the GSC model \refp{EqPstrXpt}.
\\[2mm]
{\bf Posterior Probability:} First observe that the discrete 
latent variable $\sVec$ of the GSC model can be directly combined with 
the basis functions, i.e., $W(\sVec\odot\zVec)\,=\,\Wt_{\sVec}\,\zVec,$  where $(\Wt_{\sVec})_{dh}\,=\,W_{dh}s_h$.
%
%
Now we apply the Bayes' rule to write down the posterior:
\begin{eqnarray}
p(\sVec,\zVec\,|\,\yVecN,\Theta)
&=& \frac{\NGauss(\yVecN;\,\Wt_{\sVec}\,\zVec,\Sigmad)\,\NGauss(\zVec;\,\vec{0},\Sigmah)\,p(\sVec\,|\,\Theta)}
{\sum_{\sVec^{\prime}}\int\NGauss(\yVecN;\,\Wt_{\sVec^{\prime}}\,\zVec^{\prime},\Sigmad)\,\NGauss(\zVec^{\prime};\,\vec{0},\Sigmah)\,p(\sVec^{\prime}\,|\,\Theta)\,\dz^{\prime}}.
\label{EqnPstrOrig}
\end{eqnarray}
Note that given a state $\sVec$ in \refp{EqnPstrOrig}, the Gaussian governing the 
observations $\yVecN$ is only dependent on the Gaussian over the 
continuous latent $\zVec$, which is analytically independent of $\sVec$.
We can exploit this joint relation to refactorize the Gaussians.
Using Gaussian identities the posterior can be rewritten as:
\begin{eqnarray}
p(\sVec,\zVec\,|\,\yVecN,\Theta)
&=& \frac{\NGauss(\yVecN;\vec{0},C_{\sVec})\,p(\sVec\,|\,\Theta)\,\NGauss(\zVec;\,\kappaVecN_{\sVec},\Lambda_{\sVec})}
{\sum_{\sVec^{\prime}}\NGauss(\yVecN;\vec{0},C_{\sVec^{\prime}})\,p(\sVec^{\prime}\,|\,\Theta)\,\int\NGauss(\zVec^{\prime};\,\kappaVecN_{\sVec^{\prime}},\Lambda_{\sVec^{\prime}})\,\dz^{\prime}}\nonumber\\
&=& p(\sVec\,|\,\yVecN,\Theta)\ \NGauss(\zVec;\,\kappaVecN_{\sVec},\Lambda_{\sVec}), \label{EqnPostMain} \\[-8mm]\nonumber
\label{EqnPstrRfct}
\end{eqnarray}
\begin{eqnarray}
\mbox{where}\hspace{2mm}C_{\sVec} &=& \Wt_{\sVec}\Wt^{\TT}_{\sVec}\,+\,\Sigmad,\label{EqnPostC} 
\hspace{8mm}\Lambda_{\sVec} = \big(\Wt_{\sVec}^{T}\,\Sigmad^{-1}\,\Wt_{\sVec} + \One_H\big)^{-1} \nonumber\\[1mm] 
\mbox{and}\hspace{3mm}\kappaVecN_{\sVec} &=& \Lambda_{\sVec}\,\Wt^{\TT}_{\sVec}\,\Sigmad^{-1}\,\yVecN . \label{EqnPostK} 
\end{eqnarray}
Equations \refp{EqnPostMain} to \refp{EqnPostK} represent the crucial
result for the computation of the E-step below because, first, they
show that the posterior does not involve analytically intractable
integrals and, second, for fixed $\sVec$ and $\yVecN$ the dependency
on $\zVec$ follows a Gaussian distribution. This special form allows
for the derivation of analytical expressions for the expectation
values as required for the M-step parameter updates.
\\[2mm]
{\bf E-step Equations:} Derived from \refp{EqnPostMain}, the expectation values are computed as:
\begin{eqnarray}
\E{\sVec}_n &=& \sum_{\sVec} p(\sVec\,|\,\yVecN,\Theta)\ \sVec,  \label{EqnEStepS}
\hspace{6.5mm} \E{\sVec\odot\zVec}_n = \sum_{\sVec} p(\sVec\,|\,\yVecN,\Theta)\ \kappaVecN_{\sVec} \
\label{EqnEStepSZ}\\
& &\hspace{-11mm}\mbox{and}\hspace{4mm}\E{(\sVec\odot\zVec)(\sVec\odot\zVec)^{\TT}}_n 
=
\sum_{\sVec} p(\sVec\,|\,\yVecN,\Theta)\,\big(\Lambda_{\sVec} + \kappaVecN_{\sVec}(\kappaVecN_{\sVec})^{\TT}\big).  \label{EqnEStepSZSZ}
\end{eqnarray}
Note that we have to use the current values $\Theta=\ThetaOld$ 
for all parameters on the right-hand-side. The E-step equations 
\refp{EqnEStepS} to \refp{EqnEStepSZSZ} represent
a closed-form solution for expectation values required for the
closed-form M-step \refp{EqnMStepW} to \refp{EqnMStepPi}. 
%
%
%
%
%
%
\\[2mm]
{\bf Relation to the Mixture of Gaussians:} 
The special form of the posterior in \refp{EqnPostMain} allows the derivation
of a closed-form experession of the data likelihood: i.e., 
$p(\yVec\,|\,\Theta) = \sum_{\sVec}\ \mathrm{Bernoulli}(\sVec;\piVec)\ \NGauss(\yVecN;\vec{0},C_{\sVec})$.
Note that in principle, this form 
can be reproduced by a Gaussian
mixture model. However, such a model would consist of $2^H$ mixture
components, with strongly dependent mixing proportions and 
covariance matrices $C_{\sVec}$. Closed-form
EM-updates can in general not be derived for such dependencies. The
standard updates for mixtures of Gaussians require
independently parameterized mixing proportions and components.
Therefore, the closed-form
EM-solutions for the GSC model is not a consequence of closed-form EM
for classical Gaussian mixtures.

\section{Numerical Experiments}
\begin{figure}[ht]
\vspace{-4mm}
\begin{center}
\centerline{\includegraphics[width=\columnwidth]{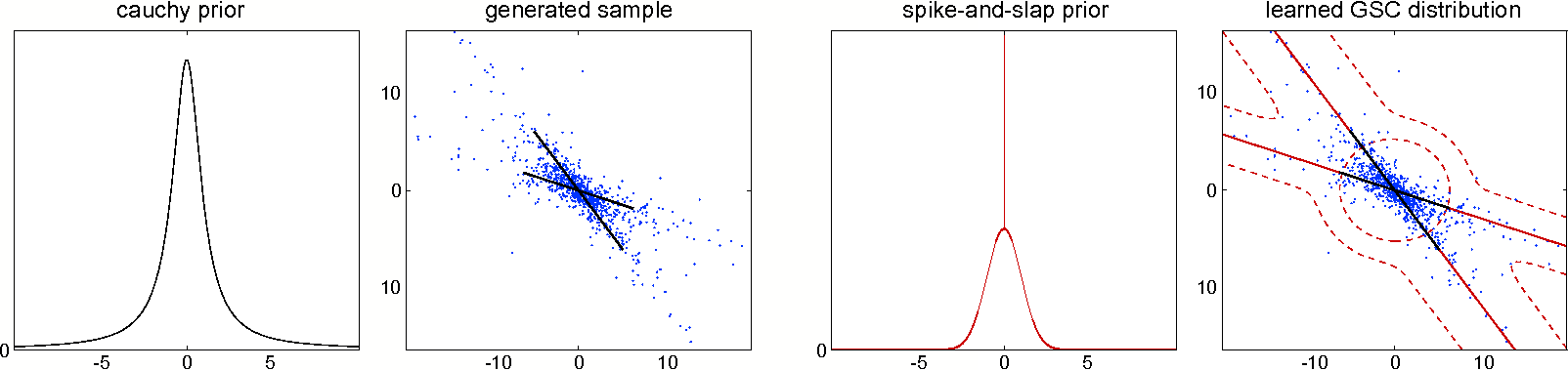}}
\vspace{-3.2mm}
\caption{Comparison of standard sparse coding and GSC.
{\bf Left panels}: Cauchy distribution (along one hidden dimension) as a standard SC prior \cite{OlshausenField1996}
 and data generated by it. 
{\bf Right panels}: Spike-and-slab distribution (one of the hidden dimensions) inferred by the GSC algorithm along with 
inferred sparse directions (solid red lines) and posterior data density contours (dotted red lines).}
%
%

\label{FigSparseDirections}
\end{center}
\vspace{-13mm}
\end{figure}

GSC parameter optimization is non-convex, 
However, as for all algorithms based on closed-form EM, 
the GSC algorithm always increases the data likelihood 
at least to a local maxima. We first numerically investigate 
how frequently local optima are obtained. Later we assess 
model's performace on more practical tasks.
\\[2mm]
{\bf Model verification:} First, we verify on artificial data that the
algorithm increases the likelihood and that it can recover the
parameters of the generating distribution. For this, we generated $N =
500$ data points $\yVecN$ from the GSC generative model
(\ref{EqnPriorS}) to (\ref{EqnNoise}) with $D = H = 2$. 
We used randomly initialized generative parameters
\footnote{We obtained
$W^{\mathrm{gen}}$ by independently drawing each matrix entry 
from a normal distribution with zero mean and
standard deviation $3$. $\pi^{\mathrm{gen}}_h$ 
values were drawn from a uniform distribution between $0.05$ and $1$, 
$\Sigmad = \sig^{\mathrm{gen}}{\One_D}$ (where $\sig^{\mathrm{gen}}$ 
was uniformly drawn between $0.05$ and $10$).}.
The algorithm was run $250$ times on the generated data.
For each run we performed $300$ EM iterations.
For each run, we randomly and uniformly initialized $\pi_h$ 
between $0.05$ and $10$, set $\Sigmad$ to the covariance accross the data points,
and the elements of $W$ we chose to be independently drawn from a normal distribution
with zero mean and unit variance. In all runs the generating parameter 
values were recovered with high accuracy. Runs with different generating 
parameters produced essentially the same results.
%
%
%
%
%
%
%
\\[2mm]
{\bf Recovery of sparse directions:} To test the model's 
robustness w.r.t.\ a relaxation of the GSC
assumptions, we applied the GSC algorithm to data generated by
standard sparse coding models. We used a standard Cauchy prior and a Gaussian 
noise model \cite{OlshausenField1996} for data generation.
Fig.\,\ref{FigSparseDirections} second panel shows data generated by this
sparse coding model while the first panel shows
the prior density along one of its hidden dimensions. We generated
$N=500$ data points with $H=D=2$. We then applied the GSC algorithm 
with the same parameter initialization as in the previous experiment. 
We performed $100$ trials using 300 EM iterations per trial. 
Again, the algorithm converged to high likelihood values in most runs (see
Fig.\,\ref{FigCauchyComp}). As a performace measure for this experiment
we investigated how well the heavy tails (i.e., the sparse directions) of
standard SC were recovered. As a performance metric, we used the 
Amari index \cite{AmariEtAl1996}:
%
%
%
\begin{figure}[t]
\begin{minipage}[b]{5cm}
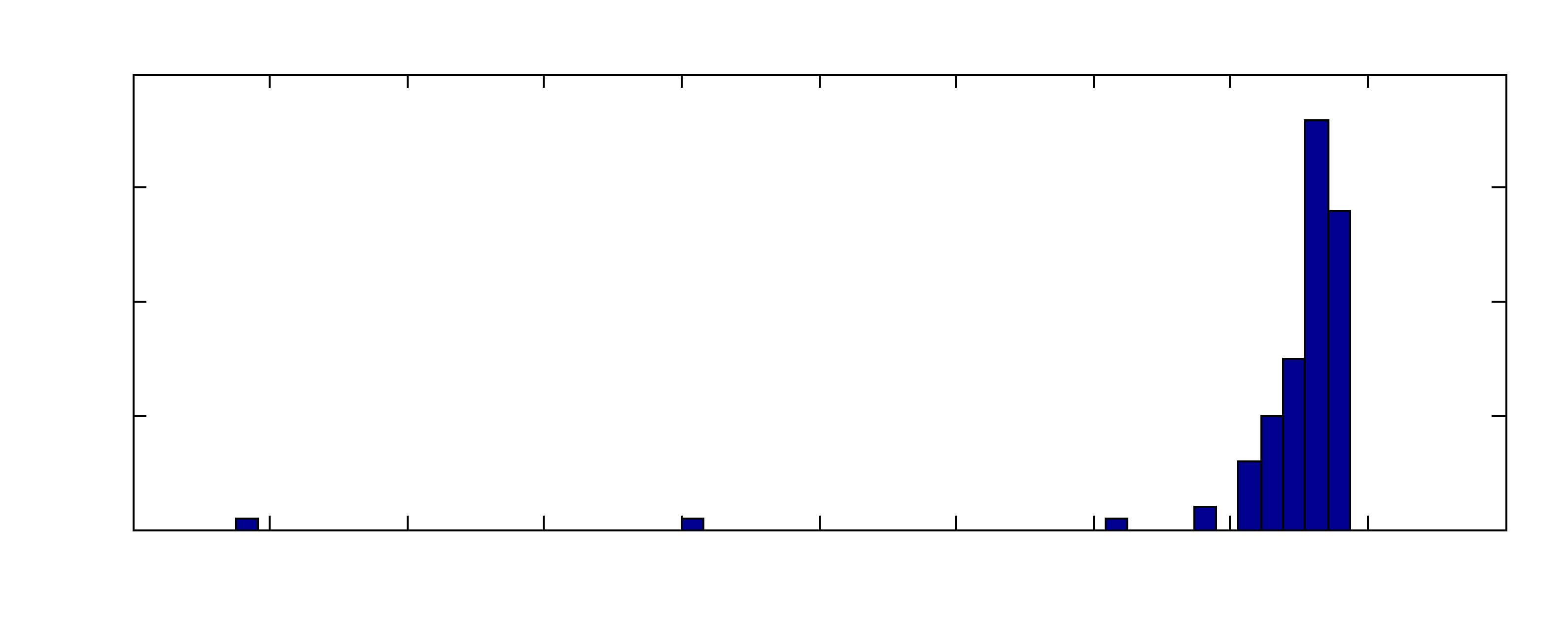
\end{minipage}\hfill
\raisebox{1.75cm}{
\begin{minipage}{4cm}
\caption{Histogram of likelihood values for 100 runs of the GSC algorithm
on data generated by a SC model with Cauchy prior. Almost all
runs converged to high likelihood values.}
\label{FigCauchyComp}
\end{minipage} }
\end{figure}
%
%
\begin{eqnarray}
\disT{}A(W) &=& \disT \frac{1}{2H(H-1)} \sum_{h,h^\prime=1}^{H} \Big(  \frac {\vert O_{hh^\prime}\vert} {\max_{h^{\prime\prime}}\vert O_{hh^{\prime\prime}}\vert}  
 +  \frac{ \vert O_{ hh^{\prime}}\vert } { \max_{h^{\prime\prime}}\vert O_{h^{\prime\prime}h^{\prime}}\vert } \Big)
- \frac{1}{H-1}
\label{EqnAmari}
\end{eqnarray}
%
where $O_{hh'}:=\left(W^{-1}W^{\mathrm{gen}}\right)_{hh'}$. 
The mean Amari index of all runs with high likelihood values was below
$10^{-2}$, which shows a very accurate recovery of the sparse
directions.  Fig.\,\ref{FigSparseDirections} (right panel) visualizes the
distribution recovered by the GSC algorithm in a typical run. The
dotted red lines show the density contours of the learned distribution
$p(\yVec\,|\,\Theta)$. High accuracy in the recovery of the
generating sparse directions (solid black lines) can be observed by
comparison with the recovered directions (solid red lines). The
results of experiments are qualitatively the same if we increase the
number of hidden and observed dimensions; e.g., for $H=D=4$ we found 
the algorithm converged to a high likelihood in $91$ (with average 
Amari index below $10^{-2}$) of $100$ runs.  
%

Other than standard SC with Cauchy prior, we also
ran the algorithm on data generated by SC with Laplace prior
\cite{OlshausenField1996,LeeEtAl2007}. There
for $H=D=2$,  we converged 
to high likelihood values in $99$ of $100$
runs with an average Amari index $0.06$. In the experiment with 
$H=D=4$ the algorithm converged to a high
likelihood in $97$ of $100$ runs. The average Amari index of
all runs with high likelihoods was $0.07$ in this case.
\\[2mm]
{\bf Source separation:} We applied the GSC algorithm to publicly
available benchmarks. We used the non-artificial benchmarks of 
\cite{SuzukiSugiyama2011}. The datasets mainly contain acoustic
data obtained from ICALAB \cite{icalab2007}. We generated the observed data by mixing 
the benchmark sources using randomly generated orthogonal
mixing matrix (we followed \cite{SuzukiSugiyama2011}).  
Again, the Amari index (\ref{EqnAmari})
was used as a performace measure.

\begin{figure}[h]
\vspace{-8mm}
\begin{minipage}[b]{6.25cm}
\centerline{\includegraphics[scale=0.41]{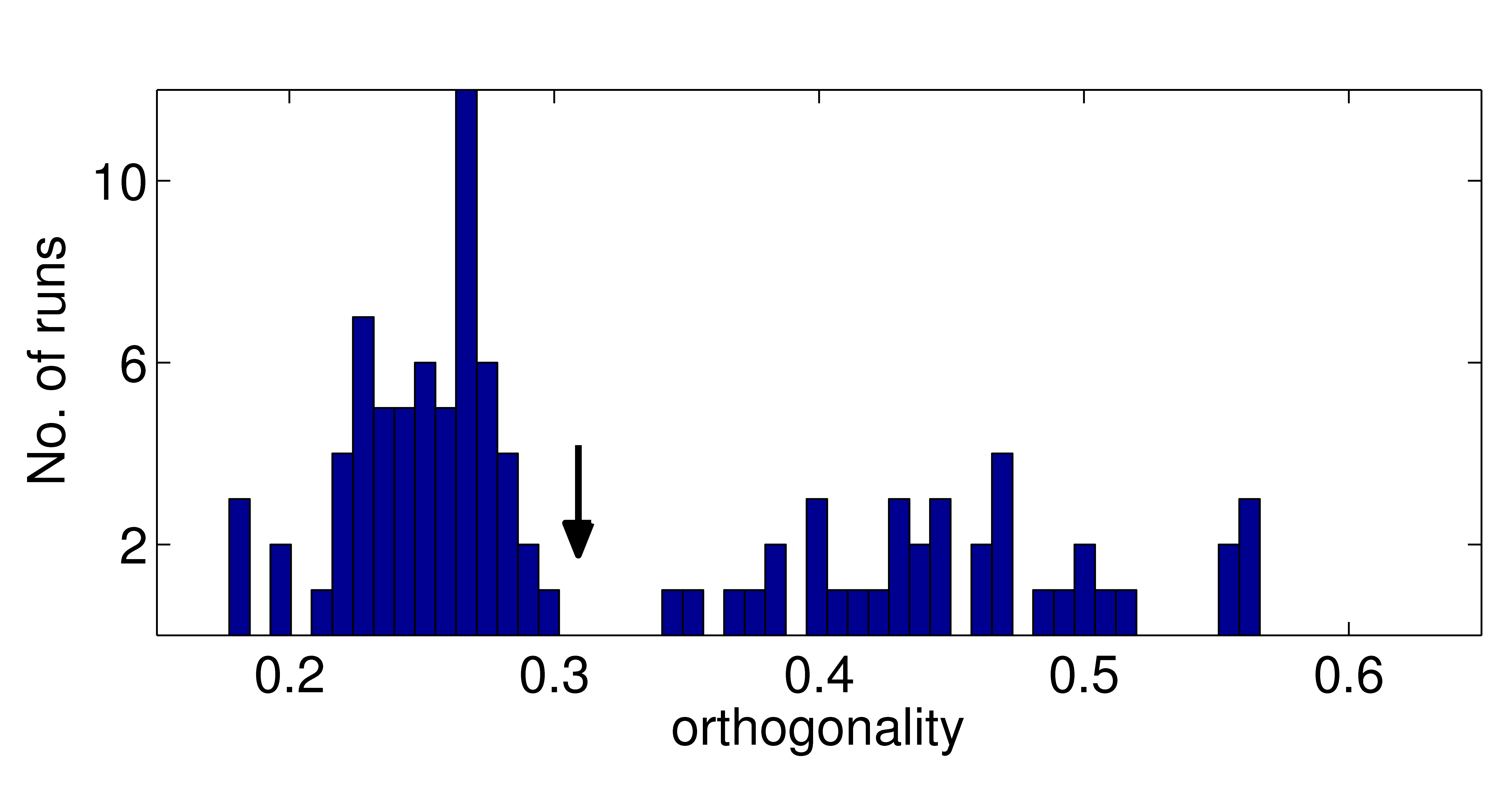}}
\end{minipage}\hfill
\raisebox{1.8cm}{
\begin{minipage}{5.5cm}
\caption{Histogram of the deviation from orthogonality of the $W$
  matrix for 100 runs of the GSC algorithm on the {\tt Speech4}
  benchmark ($N=500$). A clear cluster of the most orthogonal runs can
  automatically be detected: the threshold of runs considered is
  defined to be the minimum after the cluster (black
  arrow).
}
\label{FigSpeech4OrthoHist}
\end{minipage} }
\vspace{-6mm}
\end{figure}
\begin{table*}[t]
\footnotesize
\caption{Performance of different algorithms on benchmarks for source separation. 
Data for NG-LICA, KICA, FICA, and JADE are taken from \cite{SuzukiSugiyama2011}. 
Performances are compared based on the Amari index (\ref{EqnAmari}). Bold values 
highlight the best performing algorithm(s).}
\vspace{-1.5mm}
\label{TabLimits}
\renewcommand{\arraystretch}{1.1}
\begin{center}
\begin{tabular}{|c|c|cccccc|}\hline
\multicolumn{2}{|c|}{datasets} & \multicolumn{6}{c|}{Amari index (standard deviation)}\\\hline
name     & N & GSC & GSC$^\mathrm{\perp}$ & NG-LICA & KICA & FICA & JADE \\\hline
%
%
%
10halo   & 200 & 0.34(0.05) & \bf{0.29(0.03)} & \bf{0.29(0.02)} & 0.38(0.03) & 0.33(0.07) & 0.36(0.00)\\
         & 500 & 0.27(0.01) & 0.27(0.01) & \bf{0.22(0.02)} & 0.37(0.03) & \bf{0.22(0.03)} & 0.28(0.00)\\\hline
Sergio7  & 200 & 0.23(0.06) & 0.20(0.06) & \bf{0.04(0.01)} & 0.38(0.04) & 0.05(0.02) & 0.07(0.00)\\
         & 500 & 0.18(0.05) & 0.17(0.03) & 0.05(0.02) & 0.37(0.03) & \bf{0.04(0.01)} & \bf{0.04(0.00)}\\\hline
Speech4  & 200 & 0.25(0.05) & \bf{0.17(0.04)} & 0.18(0.03) & 0.29(0.05) & 0.20(0.03) & 0.22(0.00)\\
           & 500 & 0.11(0.04) & \bf{0.05(0.01)} & 0.07(0.00) & 0.10(0.04) & 0.10(0.04) & 0.06(0.00)\\\hline
c5signals  & 200 & 0.39(0.03) & 0.44(0.05) & 0.12(0.01) & 0.25(0.15) & \bf{0.10(0.02)} & 0.12(0.00)\\
           & 500 & 0.41(0.05) & 0.44(0.04) & 0.06(0.04) & 0.07(0.06) & \bf{0.04(0.02)} & 0.07(0.00)\\\hline
\end{tabular}
\end{center}
\vspace{-9.2mm}
\end{table*}
For all the benchmarks we used $N=200$ and $N=500$ data
points (as selected by\cite{SuzukiSugiyama2011}).  We applied GSC 
to the data using the same initialization as described
before. For each experiment we performed $100$ trials with a 
random new parameter initialization per trial. 
The first column of Tab.\,1 list average Amari indices
obtained including all trials per experiment\footnote{We 
obtained the reported results by 
diagonalizing the updated $\Sigmad$ in the M-step by setting $\Sigmad = \sig^2 \One_D$, where 
$\sig^2 = \trace(\Sigmad)/D$.}. It is important
to note that all the other algorithms listed in the comparison 
assume orthogonal mixing matrices, while the GSC algorithm does not.
Therefore in the column 'GSC$^{\perp}$' in Tab.\,1, we report 
statistics that are only computed over the runs which infered 
the most orthogonal $W$ matrices. 
As a measure of orthogonality we used the maximal deviation from $90^{o}$ between any two
axes. Fig.\,\ref{FigSpeech4OrthoHist} shows as an example a histogram of the
maximal deviations of all trials on the {\tt Speech4} data with $N=500$. As can be
observed, we obtained a clear cluster of runs with high orthogonality.
%
%
We observed worst performace of the GSC algorithm on the {\tt c5signals} 
dataset. However, the dataset contains sub-Gaussian sources 
which in general can not be recovered by sparse coding approaches.
%
%
%
%
%
%
\section{Discussion}
The GSC algorithm falls into the class of standard SC algorithms.  
However, instead of a heavy-tail prior, it uses a
spike-and-slab distribution. The algorithm has a distinguishing 
capability of taking the whole (potentially a 
multimodal) posterior into account for parameter optimization, 
which is in contrast to the MAP approximation of the posterior,
which is a widely used approach for training SC models
 (see, e.g., \cite{LeeEtAl2007,MairalBPS09})
. Various sophisticated
methods have been proposed to efficently find the MAP 
(e.g., \cite{Tibshirani1996}).\comment{here the newest paper for MAP?} 
MAP based optimizations usually also require regularization
parameters that have to be inferred (e.g., by
means of cross-validation). Other approximations that take 
more aspects of the posterior into account are also being investigated 
actively (e.g., \cite{Seeger2008,MohamedEtAl2010}).
However, approximations can introduce learning biases.
For instance, MAP and Laplace approximations assume monomodality 
in posterior estimation, which is not always the case.

Closed-form EM learning of the GSC algorithm also comes with a cost,
which is exponential w.r.t. the number of hidden 
dimensions $H$.  This can be seen by considering (\ref{EqnPstrRfct}) 
where the partition function requires
a summation over all binary vectors $\sVec$ (similar for expectation
values w.r.t.\ the posterior). 
Nevertheless, we show in numerical experiments that the algorithm is 
well applicable to the typical range of source separation tasks.  
In such domains the GSC algorithm can benefit from taking potentially
a multimodal posterior into acount and infering a whole set of
model parameters including the sparsitiy per latent dimension. For instance, when
using a number of hidden dimensions larger than the number of actual sources,
the model can discard dimensions by setting $\pi_h$ parameters to
zero. The studied model could thus be considered as treating parameter
inference in a more Bayesian way than, e.g., SC with MAP estimates (compare \cite{LeeEtAl2007}).
The second line of research aims at a fully Bayesian description of sparse coding and emphasises
a large flexibility using estimations of the number of hidden dimensions and by being applicable
with a range of different noise models. The great challenge of these general models is the procedure
of parameter estimation. For the model in \cite{MohamedEtAl2010}, for instance, the Bayesian methodology
involves conjugate priors and hyperparameters in combination with Laplace approximation
and different sampling schemes. 
While the aim, e.g.,\ in \cite{West2003,KnowlesGhahramani2007,TehGorurGhar2007,PaisleyLawrence2009,KnowlesGhahramani2010,MohamedEtAl2010} 
is a large flexibility, we aim at a generalization of
sparse coding that maintains the closed-form EM solutions.

To summarize, we have studied a novel sparse coding algorithm and have
shown its competitiveness on source separation benchmarks. Along with
the reported results on source separation, the main contribution of
this work is the derivation and numerical investigation of the (to the
knowledge of the authors) first closed-form, exact EM algorithm for
spkie-and-slab sparse coding.
\\[2mm]
{\bf Acknowledgement.} J. L\"ucke is funded by the German Research Foundation (DFG),
grant \mbox{LU 1196/4-1}; A.-S. Sheikh is funded by the German BMBF, grant 01GQ0840.\vspace{-2mm}

%
%
\bibliography{jpl2010-12Plain}
\bibliographystyle{abbrv}
\end{document}

%% file: FigGenDistributions04_cut.pstex_t
\begin{picture}(0,0)%
\includegraphics{FigGenDistributions04_cut.png}%
\end{picture}%
\setlength{\unitlength}{1865sp}%
\begingroup\makeatletter\ifx\SetFigFont\undefined%
\gdef\SetFigFont#1#2#3#4#5{%
  \reset@font\fontsize{#1}{#2pt}%
  \fontfamily{#3}\fontseries{#4}\fontshape{#5}%
  \selectfont}%
\fi\endgroup%
\begin{picture}(7693,5282)(2431,-7728)
\put(2814,-6503){\makebox(0,0)[rb]{\smash{{\SetFigFont{7}{8.4}{\rmdefault}{\mddefault}{\updefault}{\color[rgb]{0,0,0}$0$}%
}}}}
\put(2814,-5532){\makebox(0,0)[rb]{\smash{{\SetFigFont{7}{8.4}{\rmdefault}{\mddefault}{\updefault}{\color[rgb]{0,0,0}$5$}%
}}}}
\put(2814,-7466){\makebox(0,0)[rb]{\smash{{\SetFigFont{7}{8.4}{\rmdefault}{\mddefault}{\updefault}{\color[rgb]{0,0,0}$-5$}%
}}}}
\put(5273,-6503){\makebox(0,0)[rb]{\smash{{\SetFigFont{7}{8.4}{\rmdefault}{\mddefault}{\updefault}{\color[rgb]{0,0,0}$0$}%
}}}}
\put(5273,-5532){\makebox(0,0)[rb]{\smash{{\SetFigFont{7}{8.4}{\rmdefault}{\mddefault}{\updefault}{\color[rgb]{0,0,0}$5$}%
}}}}
\put(5273,-7466){\makebox(0,0)[rb]{\smash{{\SetFigFont{7}{8.4}{\rmdefault}{\mddefault}{\updefault}{\color[rgb]{0,0,0}$-5$}%
}}}}
\put(7727,-6503){\makebox(0,0)[rb]{\smash{{\SetFigFont{7}{8.4}{\rmdefault}{\mddefault}{\updefault}{\color[rgb]{0,0,0}$0$}%
}}}}
\put(7727,-5532){\makebox(0,0)[rb]{\smash{{\SetFigFont{7}{8.4}{\rmdefault}{\mddefault}{\updefault}{\color[rgb]{0,0,0}$5$}%
}}}}
\put(7727,-7466){\makebox(0,0)[rb]{\smash{{\SetFigFont{7}{8.4}{\rmdefault}{\mddefault}{\updefault}{\color[rgb]{0,0,0}$-5$}%
}}}}
\put(3830,-5125){\makebox(0,0)[b]{\smash{{\SetFigFont{7}{8.4}{\rmdefault}{\mddefault}{\updefault}{\color[rgb]{0,0,0}$0$}%
}}}}
\put(4771,-5125){\makebox(0,0)[b]{\smash{{\SetFigFont{7}{8.4}{\rmdefault}{\mddefault}{\updefault}{\color[rgb]{0,0,0}$5$}%
}}}}
\put(2881,-5125){\makebox(0,0)[b]{\smash{{\SetFigFont{7}{8.4}{\rmdefault}{\mddefault}{\updefault}{\color[rgb]{0,0,0}$-5$}%
}}}}
\put(6289,-5125){\makebox(0,0)[b]{\smash{{\SetFigFont{7}{8.4}{\rmdefault}{\mddefault}{\updefault}{\color[rgb]{0,0,0}$0$}%
}}}}
\put(7230,-5125){\makebox(0,0)[b]{\smash{{\SetFigFont{7}{8.4}{\rmdefault}{\mddefault}{\updefault}{\color[rgb]{0,0,0}$5$}%
}}}}
\put(5340,-5125){\makebox(0,0)[b]{\smash{{\SetFigFont{7}{8.4}{\rmdefault}{\mddefault}{\updefault}{\color[rgb]{0,0,0}$-5$}%
}}}}
\put(8753,-5125){\makebox(0,0)[b]{\smash{{\SetFigFont{7}{8.4}{\rmdefault}{\mddefault}{\updefault}{\color[rgb]{0,0,0}$0$}%
}}}}
\put(9694,-5125){\makebox(0,0)[b]{\smash{{\SetFigFont{7}{8.4}{\rmdefault}{\mddefault}{\updefault}{\color[rgb]{0,0,0}$5$}%
}}}}
\put(7804,-5125){\makebox(0,0)[b]{\smash{{\SetFigFont{7}{8.4}{\rmdefault}{\mddefault}{\updefault}{\color[rgb]{0,0,0}$-5$}%
}}}}
\put(5273,-3976){\makebox(0,0)[rb]{\smash{{\SetFigFont{7}{8.4}{\rmdefault}{\mddefault}{\updefault}{\color[rgb]{0,0,0}$0$}%
}}}}
\put(5273,-3005){\makebox(0,0)[rb]{\smash{{\SetFigFont{7}{8.4}{\rmdefault}{\mddefault}{\updefault}{\color[rgb]{0,0,0}$5$}%
}}}}
\put(5273,-4939){\makebox(0,0)[rb]{\smash{{\SetFigFont{7}{8.4}{\rmdefault}{\mddefault}{\updefault}{\color[rgb]{0,0,0}$-5$}%
}}}}
\put(7727,-3976){\makebox(0,0)[rb]{\smash{{\SetFigFont{7}{8.4}{\rmdefault}{\mddefault}{\updefault}{\color[rgb]{0,0,0}$0$}%
}}}}
\put(7727,-3005){\makebox(0,0)[rb]{\smash{{\SetFigFont{7}{8.4}{\rmdefault}{\mddefault}{\updefault}{\color[rgb]{0,0,0}$5$}%
}}}}
\put(7727,-4939){\makebox(0,0)[rb]{\smash{{\SetFigFont{7}{8.4}{\rmdefault}{\mddefault}{\updefault}{\color[rgb]{0,0,0}$-5$}%
}}}}
\put(3830,-7655){\makebox(0,0)[b]{\smash{{\SetFigFont{7}{8.4}{\rmdefault}{\mddefault}{\updefault}{\color[rgb]{0,0,0}$0$}%
}}}}
\put(4771,-7655){\makebox(0,0)[b]{\smash{{\SetFigFont{7}{8.4}{\rmdefault}{\mddefault}{\updefault}{\color[rgb]{0,0,0}$5$}%
}}}}
\put(2881,-7655){\makebox(0,0)[b]{\smash{{\SetFigFont{7}{8.4}{\rmdefault}{\mddefault}{\updefault}{\color[rgb]{0,0,0}$-5$}%
}}}}
\put(6289,-7655){\makebox(0,0)[b]{\smash{{\SetFigFont{7}{8.4}{\rmdefault}{\mddefault}{\updefault}{\color[rgb]{0,0,0}$0$}%
}}}}
\put(7230,-7655){\makebox(0,0)[b]{\smash{{\SetFigFont{7}{8.4}{\rmdefault}{\mddefault}{\updefault}{\color[rgb]{0,0,0}$5$}%
}}}}
\put(5340,-7655){\makebox(0,0)[b]{\smash{{\SetFigFont{7}{8.4}{\rmdefault}{\mddefault}{\updefault}{\color[rgb]{0,0,0}$-5$}%
}}}}
\put(8753,-7655){\makebox(0,0)[b]{\smash{{\SetFigFont{7}{8.4}{\rmdefault}{\mddefault}{\updefault}{\color[rgb]{0,0,0}$0$}%
}}}}
\put(9694,-7655){\makebox(0,0)[b]{\smash{{\SetFigFont{7}{8.4}{\rmdefault}{\mddefault}{\updefault}{\color[rgb]{0,0,0}$5$}%
}}}}
\put(7804,-7655){\makebox(0,0)[b]{\smash{{\SetFigFont{7}{8.4}{\rmdefault}{\mddefault}{\updefault}{\color[rgb]{0,0,0}$-5$}%
}}}}
\put(2814,-3976){\makebox(0,0)[rb]{\smash{{\SetFigFont{7}{8.4}{\rmdefault}{\mddefault}{\updefault}{\color[rgb]{0,0,0}$0$}%
}}}}
\put(2814,-3005){\makebox(0,0)[rb]{\smash{{\SetFigFont{7}{8.4}{\rmdefault}{\mddefault}{\updefault}{\color[rgb]{0,0,0}$5$}%
}}}}
\put(2814,-4939){\makebox(0,0)[rb]{\smash{{\SetFigFont{7}{8.4}{\rmdefault}{\mddefault}{\updefault}{\color[rgb]{0,0,0}$-5$}%
}}}}
\put(6294,-5316){\makebox(0,0)[b]{\smash{{\SetFigFont{7}{8.4}{\rmdefault}{\mddefault}{\updefault}{\color[rgb]{0,0,0}{\sf observed samples}}%
}}}}
\put(6294,-2792){\makebox(0,0)[b]{\smash{{\SetFigFont{7}{8.4}{\rmdefault}{\mddefault}{\updefault}{\color[rgb]{0,0,0}{\sf prior samples}}%
}}}}
\put(4363,-3236){\makebox(0,0)[b]{\smash{{\SetFigFont{7}{8.4}{\rmdefault}{\mddefault}{\updefault}{\color[rgb]{0,0,0}$\pi_h$=$1$}%
}}}}
\put(6817,-3236){\makebox(0,0)[b]{\smash{{\SetFigFont{7}{8.4}{\rmdefault}{\mddefault}{\updefault}{\color[rgb]{0,0,0}$\pi_h$=$\frac{1}{2}$}%
}}}}
\put(9275,-3236){\makebox(0,0)[b]{\smash{{\SetFigFont{7}{8.4}{\rmdefault}{\mddefault}{\updefault}{\color[rgb]{0,0,0}$\pi_h$=$\frac{1}{4}$}%
}}}}
\end{picture}%

%% file: FigCauchyComp02.pstex_t
\begin{picture}(0,0)%
\includegraphics{FigCauchyComp02.png}%
\end{picture}%
\setlength{\unitlength}{1326sp}%
\begingroup\makeatletter\ifx\SetFigFont\undefined%
\gdef\SetFigFont#1#2#3#4#5{%
  \reset@font\fontsize{#1}{#2pt}%
  \fontfamily{#3}\fontseries{#4}\fontshape{#5}%
  \selectfont}%
\fi\endgroup%
\begin{picture}(11348,4557)(1420,-7003)
\put(4231,-6676){\makebox(0,0)[b]{\smash{{\SetFigFont{8}{9.6}{\rmdefault}{\mddefault}{\updefault}{\color[rgb]{0,0,0}$-5.5$}%
}}}}
\put(8236,-6676){\makebox(0,0)[b]{\smash{{\SetFigFont{8}{9.6}{\rmdefault}{\mddefault}{\updefault}{\color[rgb]{0,0,0}$-5.3$}%
}}}}
\put(10216,-6676){\makebox(0,0)[b]{\smash{{\SetFigFont{8}{9.6}{\rmdefault}{\mddefault}{\updefault}{\color[rgb]{0,0,0}$-5.2$}%
}}}}
\put(6211,-6676){\makebox(0,0)[b]{\smash{{\SetFigFont{8}{9.6}{\rmdefault}{\mddefault}{\updefault}{\color[rgb]{0,0,0}$-5.4$}%
}}}}
\put(2341,-6406){\makebox(0,0)[rb]{\smash{{\SetFigFont{8}{9.6}{\rmdefault}{\mddefault}{\updefault}{\color[rgb]{0,0,0}$0$}%
}}}}
\put(2341,-5596){\makebox(0,0)[rb]{\smash{{\SetFigFont{8}{9.6}{\rmdefault}{\mddefault}{\updefault}{\color[rgb]{0,0,0}$10$}%
}}}}
\put(2341,-4786){\makebox(0,0)[rb]{\smash{{\SetFigFont{8}{9.6}{\rmdefault}{\mddefault}{\updefault}{\color[rgb]{0,0,0}$20$}%
}}}}
\put(2341,-3976){\makebox(0,0)[rb]{\smash{{\SetFigFont{8}{9.6}{\rmdefault}{\mddefault}{\updefault}{\color[rgb]{0,0,0}$30$}%
}}}}
\put(12331,-6676){\makebox(0,0)[b]{\smash{{\SetFigFont{8}{9.6}{\rmdefault}{\mddefault}{\updefault}{\color[rgb]{0,0,0}$\log({\cal L}(\Theta))$}%
}}}}
\put(1711,-4696){\rotatebox{90.0}{\makebox(0,0)[b]{\smash{{\SetFigFont{8}{9.6}{\rmdefault}{\mddefault}{\updefault}{\color[rgb]{0,0,0}{\sf No of runs}}%
}}}}}
\end{picture}%